\documentclass[sigconf]{acmart}
 \def\BibTeX{{\rm B\kern-.05em{\sc i\kern-.025em b}\kern-.08emT\kern-.1667em\lower.7ex\hbox{E}\kern-.125emX}}

\usepackage{multirow}
\usepackage{mwe}
\usepackage{tikz}
\usepackage{booktabs}
\begin{document}

%
% The "title" command has an optional parameter, allowing the author to define a "short title" to be used in page headers.
\title{ParNet: Position-aware Aggregated Relation Network for Image-Text matching}

% The "author" command and its associated commands are used to define the authors and their affiliations.
% Of note is the shared affiliation of the first two authors, and the "authornote" and "authornotemark" commands
% used to denote shared contribution to the research.
\author{Yaxian Xia}
\affiliation{%
  \institution{Peking University}
  \streetaddress{2199 Lishui Rd}
  \city{Nanshan District}
  \state{Shenzhen}
  \country{China}}
\email{xiayaxian@pku.edu.cn}

\author{Lun Huang}
\affiliation{%
  \institution{Peking University}
  \streetaddress{2199 Lishui Rd}
  \city{Nanshan District}
  \state{Shenzhen}
  \country{China}}
\email{huanglun@pku.edu.cn}

\author{Wenmin Wang}
\affiliation{%
  \institution{Peking University}
  \streetaddress{2199 Lishui Rd}
  \city{Nanshan District}
  \state{Shenzhen}
  \country{China}}
\email{wangwm@ece.pku.edu.cn}

\author{Xiaoyong Wei}
\affiliation{%
  \institution{Pengcheng Laboratory}
  \streetaddress{xxxx}
  \city{Nanshan District}
  \state{Shenzhen}
  \country{China}}
\email{weixy@pcl.ac.cn}

\author{Jie Chen}
\affiliation{%
  \institution{Pengcheng Laboratory}
  \streetaddress{xxxx}
  \city{Nanshan District}
  \state{Shenzhen}
  \country{China}}
\email{chenj@pcl.ac.cn}

%
% By default, the full list of authors will be used in the page headers. Often, this list is too long, and will overlap
% other information printed in the page headers. This command allows the author to define a more concise list
% of authors' names for this purpose.

%
% The abstract is a short summary of the work to be presented in the article.
\begin{abstract}
Exploring fine-grained relationship between entities(e.g. objects in image or words in sentence) has great contribution to understand multimedia content precisely.
Previous attention mechanism employed in image-text matching either takes multiple self attention steps to gather correspondences or uses image objects (or words) as context to infer image-text similarity. However, they only take advantage of semantic information without considering that objects' relative position also contributes to image understanding.
To this end, we introduce a novel position-aware relation module to model both the semantic and spatial relationship simultaneously for image-text matching in this paper.
Given an image, our method utilizes the location of different objects to capture spatial relationship innovatively. With the combination of semantic and spatial relationship, it's easier to understand the content of different modalities (images and sentences) and capture fine-grained latent correspondences of image-text pairs.
Besides, we employ a two-step aggregated relation module to capture interpretable alignment of image-text pairs.
The first step, we call it intra-modal relation mechanism, in which we computes responses between different objects in an image or different words in a sentence separately; The second step, we call it inter-modal relation mechanism, in which the query plays a role of textual context to refine the relationship among object proposals in an image.
In this way, our position-aware aggregated relation network (ParNet) not only knows which entities are relevant by attending on different objects (words) adaptively, but also adjust the inter-modal correspondence according to the latent alignments according to query's content.  Our approach achieves the state-of-the-art results on MS-COCO dataset.

\end{abstract}


%\ccsdesc[500]{Information systems ~Multimedia and multimodal retrieval}
% Keywords. The author(s) should pick words that accurately describe the work being
% presented. Separate the keywords with commas.
\maketitle
\section{Introduction}
Relationship exploring between image and natural language is of central importance to various tasks, such as image-text matching, image caption and visual question answering (VQA).
\begin{figure}[t]
\begin{center}
\includegraphics[width=0.99\linewidth]{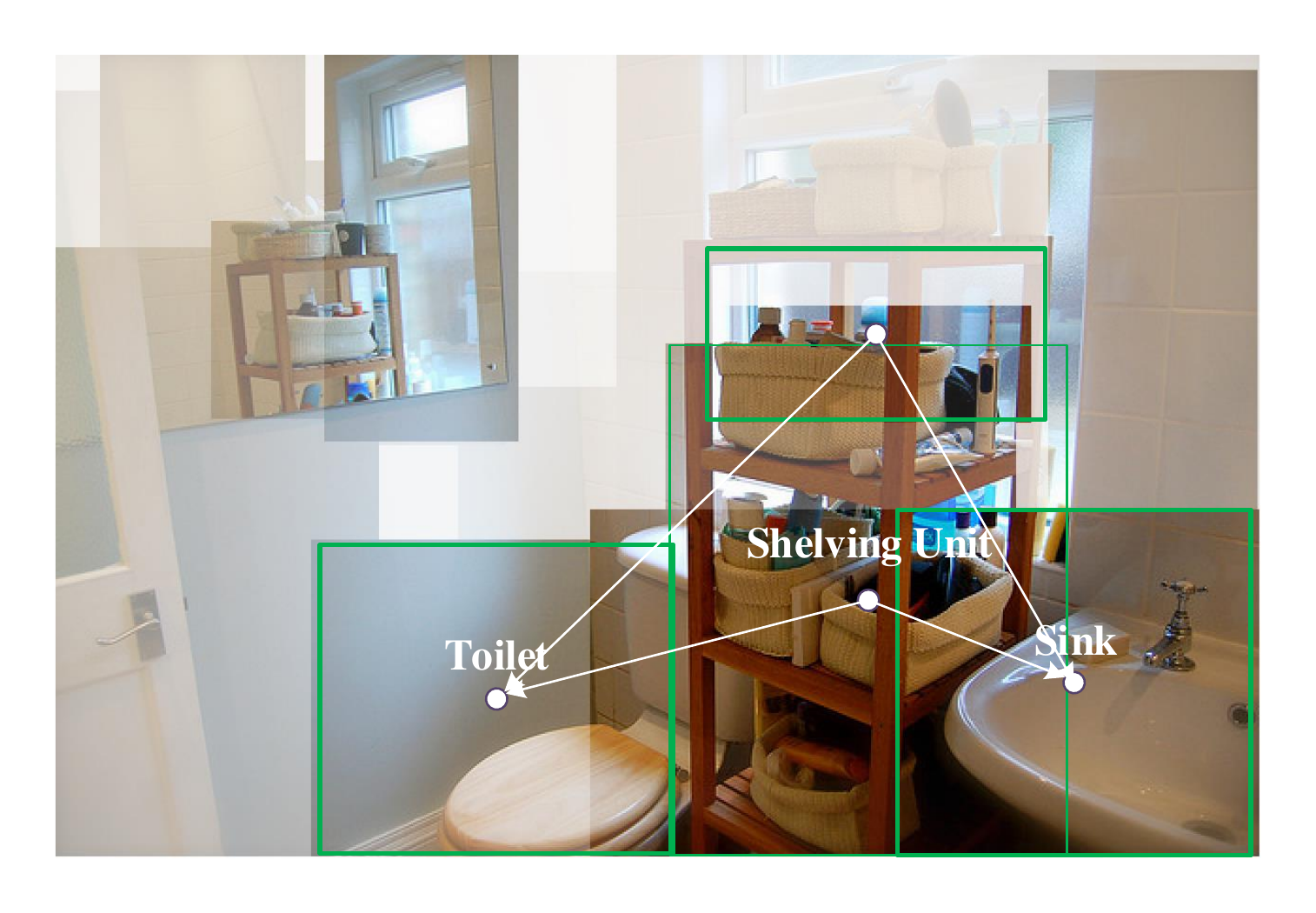}
\end{center}
   \caption{ An image with description "A shelving unit is \textbf{\textit{between}} a toilet and a bathroom sink". Sentence description makes reference to both objects (\textit{e.g.}, 'toilet', 'shelving unit' and 'sink' ) and relative position of objects (\textit{e.g.,between }).  Identifying spatial relation of objects contributes to image-text matching.}
\label{fig:long}
\label{fig:onecol}
\end{figure}

Image-text matching task requires a combination of concepts from computer version and Natural Language Processing. Methods for image-text matching are required to understand the contents of both sentence and image correctly as well as modelling their interactions. For the existence of heterogeneity gap between different media types, learning intrinsic correspondences of image-text pair is quite challenging.

Information distributions of different modalities are really inconsistent. For instance, text contains abundant prepositions to express locative and spatial relationship.
As shown in Fig.1, consider an example sentence, such as "A shelving unit is \textbf{\textit{between}} a toilet and a bathroom sink". In order to retrieve a corresponding image correctly, all entities are required to be identified precisely as well as relationships present in the sentence (\textit{between}). Although prepositions in a sentence, \textit{e.g.,} \textit{in}, \textit{between} and \textit{on},  correspond to none of the entities in this image, they influence the relevance among different image objects quite a lot. For instance, 'A shelving unit is between a toilet and a sink' and  'A toilet is between a shelving unit and a sink' contain the same objects, but correspond to different image sceneries. Position relationship is crucial to understand such rich image content. So it is crucial to model semantic and spatial relationships between object features effectively and efficiently.

Some pioneering methods focus on objects of image and phase in sentences separately to capture fine-level alignments than those methods that directly map a whole image or sentence into a common representation space. Karpathy \textit{et al.} \cite{KarpathyJF14} broke down and embedded objects in an image and phrases in a sentence into a common embedding space and calculated the similarity of object-phrase pairs. However, an image is described as a set of object representation without location.

Most of methods mentioned above ignored the spatial position of image objects, which is important to correctly understand image descriptions (e.g. 'A \textit{shelving unit} is between a \textit{toliet} and a \textit{bathroom sink}' means totally different to 'A \textit{toliet} is between a \textit{shelving unit} and a \textit{bathroom sink}').
Interaction and spatial relationship among objects are almost ignored. And the effect of spatial relation between objects in image is suppressed. Objects in image are isolated spatially.  It is difficult to obtain spatial relationships described in corresponding words.

In our proposed approach, a position-aware module is employed to capture both relative position information and semantic information for image. Our method utilizes location of different objects in an image to capture geometric distributions innovatively. With the combination of semantic and spatial information, it's easy to understand the content of different modalities and capture interpretable latent correspondences of image-text pairs.

Besides, Attention mechanisms have achieved great success in Neural Machine Translation. It has the strength to model the dependencies between relevant aspects of data without regard to their distances. Motivated by the success of attention mechanisms in NLP, several state-of-the-art approaches \cite{Nam_2017_CVPR} \cite{Lee_2018_ECCV} have improved the performance by employing attention mechanisms to capture the latent semantic correspondences between object-word pairs.

Inspired by the success of attention mechanism, we aggregate intra-modal relation and inter-modal relation into a two-step relation module.

The first step, we call it intra-modal relation mechanism, in which we computes responses between different objects in an image or different words in a sentence separately and attend on informative parts within each modality. Intra-modal relation between objects contribute to fine content understanding in a certain image or a sentence with a sense of global consciousness.

The second step, we call it inter-modal relation mechanism, in which the query plays a role of textual context to refine the relationship among object proposals in an image.
 Note that, given different queries, not all entities are equally informative. The relevance of the modality and query's intent is evaluated adaptively.

The aggregation of two-way relation mechanism is expected to perform better than one.
In this way, our aggregated relation mechanism not only know which entities are relevant by attending on different objects (words) adaptively, but also adjust the inter-modal relationship according to the latent alignments with corresponding sentence (image).

To sum up, Figure 2 illustrates the structure of position-aware aggregated relation network (ParNet). There are three parts of our proposed method: intra-modal relation network with position-aware module for image, intra-modal relation network for text and inter-modal relation network. Our main contribution are as follows:
\begin{itemize}
\item We propose a position-aware relation module to  capture spatial and semantic relationships simultaneous among objects for image.
\item A two-step relation mechanism is aggregated in our method. First, correlation among different entities of the inputs (image and text) are explored for each modality seperately; second, an image (sentence) search system calculates the importance of each modality according to different queries.
\item Extensive experiments validates our approach. Our ParNet achieves the state-of-the-art performance on image-text matching tasks on MS-COCO dataset.
\end{itemize}
\begin{figure*}
\begin{center}
\includegraphics[width=0.95\textwidth]{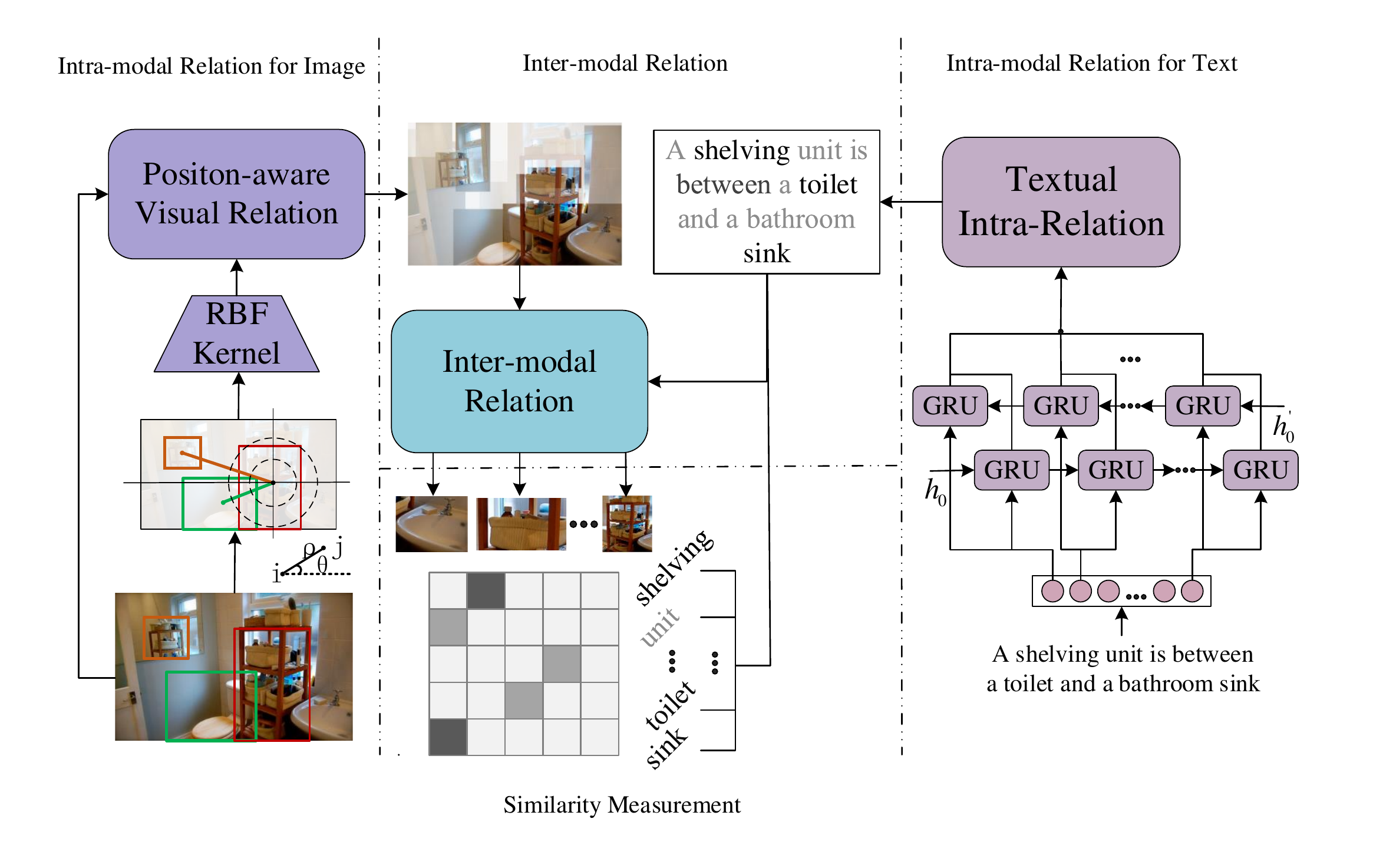}
%\fbox{\rule{0pt}{2in} \rule{.9\linewidth}{0pt}}
\end{center}
   \caption{Overview of our two-step aggregated relation network with position-aware module.
The first step, intra-modal relation mechanism, computes responses between different entities in an image or a sentence separately, a position-aware relation module is employed for image to capture semantic and spatial relation simultaneously; In the second step, inter-modal relation mechanism, text plays a role of textual weak  annotation to contribute to image understanding with corresponding description.}
\label{fig:short}
\end{figure*}

\section{Related works}
\subsection{Attention Mechanisms}
Attention mechanism have recently been successfully applied in the Natural Language Processing field\cite{NIPS2017_7181}\cite{DBLP:journals/corr/abs-1807-03819}\cite{DBLP:journals/corr/abs-1810-04805}.
The attention module can well capture the long-term dependencies at a position which allow models to automatically focus on important parts of inputs. The strong strength of it is the ability to parallel implementation. In NLP field, there is a recent trend of replacing recurrent neural networks by attention models. Transformer\cite{NIPS2017_7181} proposed multi-head attention module, which outperforms than state-of-the-art methods in Neural Machine Translation. It has the ability to focus on different relation from different heads at different words.

A self-attention mechanism computes the importance between a sequence of inputs (\textit{e.g.}, a set of words or a set of object proposals) by attending to all elements and returns a weighted vector in a representation space.
There are two kinds of widely used functions, additive attention and dot-product attention. Additive attention computes the dependence function through a feed-forward layer.
Theoretically, the complexity of these two methods are similar, but dot-product attention is more time and space-efficient in practice.

Recently,attention mechanisms have been successfully migrated to optimize object detector \cite{Hu_2018_CVPR}\cite{DBLP:journals/corr/abs-1711-07971} and understand multimedia content, \textit{e.g.}, image captioning, cross-modal retrieval \cite{Lee_2018_ECCV}\cite{Nam_2017_CVPR} and visual question answering. Nam \textit{et al.}\cite{Nam_2017_CVPR} proposed DAN to capture fine-grained interplay between vision and language through multiple step self-attention.

Inspired by the success of attention mechanism, we propose a two-step aggregated relation network with position-aware module for image-text matching, which can explore intra-modal and inter-modal relationship simultaneously.
\subsection{Image-Text Matching}
The core challenge of image-text matching is that is that data from different modalities have inconsistent semantic distributions
and learning the intrinsic correlation between them is quite elusive.
It is crucial to find accurate and fine-grained correspondence for image-text pairs.

The mainstream methods for this task are learning a comparable representations for images and sentences in a single subspace\cite{NIPS2013_5204}\cite{DBLP:journals/corr/TsaiHS17}.
Canonical correlation analysis(CCA)\cite{Hardoon2004Canonical} is employed by Hardoon \textit{et al.} to learn a common space to maximize correlation between query and image.
Word2VisualVec\cite{Dong2016Word2VisualVec} learned to predict visual representations of textual items based on a DNN architecture.
Andrej \textit{et al.} \cite{KarpathyJF14}, embed images objects and fragments of sentences into a common space to explore object-level alignment.
Bokun \textit{et al.} \cite{Wang2017Adversarial}, proposed ACMR which is built around the concept of adversarial learning.
Nam \textit{et al.} \cite{Nam_2017_CVPR}, takes multiple self attention steps to gather correspondences.
Lee \textit{et al.} \cite{Lee_2018_ECCV}, uses image objects (or words) as context to infer image-text similarity.
To the best of our knowledge, few study has attempted to explore spatial relationship. Quite a lot of position information are abundant while processing object features. And the attention mechanism they employed for cross modal retrieval are  not flexible enough.
\subsection{Kernel functions}
Kernel functions transform data into another dimension that has a clear margin between classes of data\cite{Monti_2017_CVPR}\cite{Scholkopf:2001:LKS:559923}.
So that low-dimensional data is enable to be classified in a high dimension, while the explicit mapping function is not necessary.
The most popular kernel function is radial basis function kernel (RBF kernel). It is commonly used in support vector machine classification.
% The definition of RBF kernel is as follows:
% \begin{equation}\label{equa:2}
% \begin{aligned}
% K(x_1,x_2)=exp(-\frac{\parallel x_1-x_2 \parallel^2}{2\sigma^2})
% \end{aligned} \end{equation}
It has been proved to be effective by previous work. In our method, RBF kernels are employed to map our polar coordinates of objects into higher separable.

\section{Position-aware Aggregated Relation Network}
On significant goal of cross-modal retrieval is to learn a comparable common space for image-text pair descriptions. In this paper, we propose a two-step relation network with a position-aware module for image-text matching. This module is used to capturing both the fully latent semantic and spatial object relations. In this section, our proposed position-aware aggregated relation network (ParNet) is described in detail.

On the one hand, our method extracts a set of word representations through word embedding and bi-directional Gated Recurrent Units (GRUs)\cite{650093}. On the other hand, a set of image object descriptions consisting of bounding box coordinates and object features generated by object-detector (Faster-RCNN\cite{DBLP:journals/corr/RenHG015}).

Two-step aggregated relation module is then constructed to capture the intra-modal and inter-modal relationships simultaneously. The intra-modal relation module is employed to capture query-unrelated relationships within each modality and attend on the most informative parts for each modality(As illustrated in Figure 1, object 'sink' is relevant to object 'toilet' and 'shelving unit' under the in the surroundings of bath room.) Subsequently, given different queries, information contributions of different modalities are various. A retrieval system should pay attention to balance the importance of each modality according to query's intent (inter-modal relation). Notably, relative position of different image objects are considered in our method. Latent semantic and spatial relationship are explored for image.

The image-text similarity is measured by aligning object-word pairs .

%-------------------------------------------------------------------------
\subsection{Input Representation}
We describe the embedding vectors computed for both the input image and sentence in this subsection.
\subsubsection{Image Representations}
For images, a set of object proposals are detected to represent for an image. The corresponding area of the convolutional feature map is extracted by object detector for each proposed bounding box.

In particular, we detect bottom-up attention features corresponding to salient image objects in each image with a Faster-RCNN\cite{DBLP:journals/corr/RenHG015}, which is pretrained by Anderson \textit{et al.}\cite{Anderson_2018_CVPR} on Visual Genome\cite{DBLP:journals/corr/KrishnaZGJHKCKL16}. Each object in Faster-RCNN focuses on a complete object while the traditional CNN (\textit{e.g.},VGG\cite{article}) produces a grid of features without clear boundary. Besides, coordinates of bounding box are easy to obtain through Faster-RCNN.

Image is represented as $I=\{i_1,i_2,...,i_N\}$, where $i_k$ encodes k-th detected object in the image (bounding box coordinates and associated feature vector) and $N$ is the number of image objects. $i_k$ is concatenatd by 2048-dimension object features $v_k$ and 4-dimension bounding box coordinates  $p_k$. The whole image is writed as $I=[V||P]=\{v_1||p_1,v_2||p_2,...,v_n||p_n\}$.
\subsubsection{Text Representations}
Text features are generated from bidirectional Gated Recurrent Unit(GRU) as depicted in Fig.2. First, word embeddings are employed to convert the one-hot encoding of an input sentence into a variable-dimension vector space. Then we feed the sequence of wording embeddings into bidirectional GRUs to construct a set of word vectors.

GRU can keep track of arbitrary long-term dependencies in the input sequences without the vanishing gradient problem, which performs like a long short-term memory (LSTM) but has fewer parameters than LSTM. And GRU is faster to train without losing any precision. Text is described as $T=\{t_1,t_2,...,t_M\}$, where $M$ is the length of sentence. $t_k$ encodes the representation of the k-th word in the context of the entire sentences.

\subsection{Position-aware relation module for image}
As shown in Fig.3, a position-aware relation module is proposed for capturing semantic and spatial object-level relation simultaneously. It aims to enrich the expression of objects by adaptively focusing on both spatial-relevant and semantic-relevant parts of the input image.

\begin{figure*}[t]
\begin{center}
\includegraphics[width=0.80\textwidth]{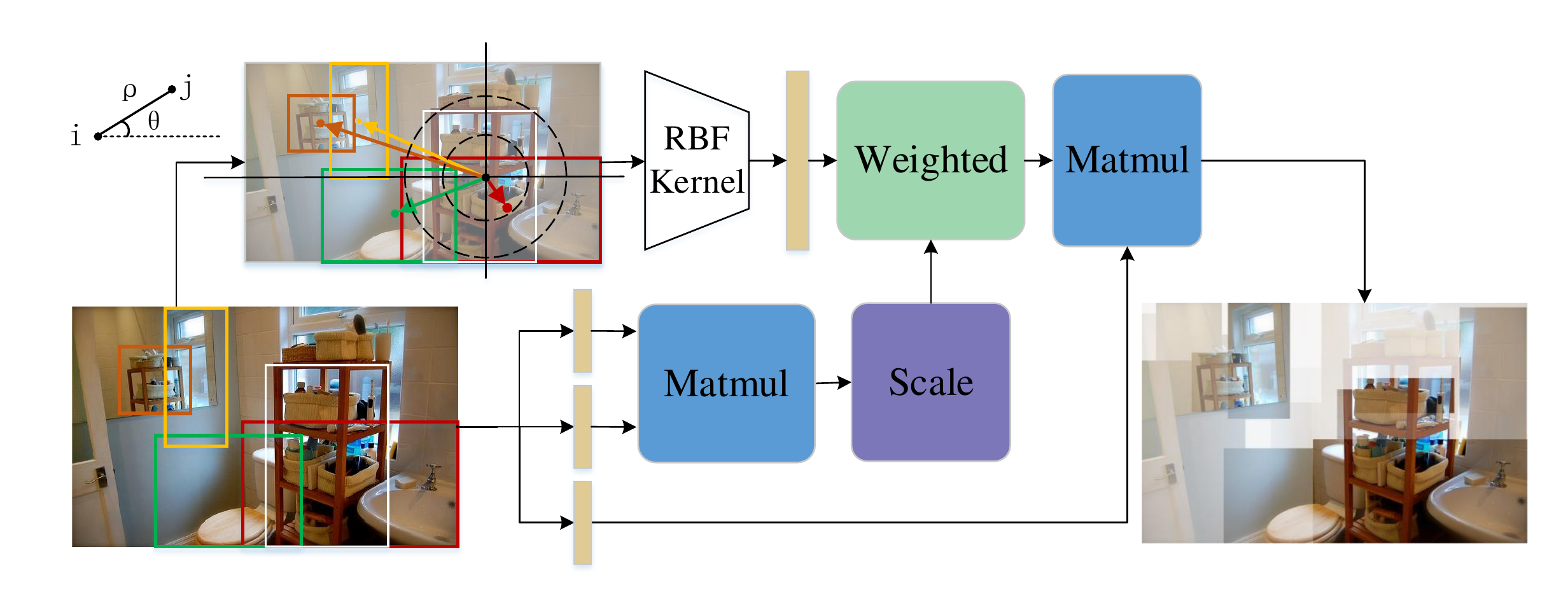}
\end{center}
   \caption{Position-aware relation module for image }
\label{fig:long}
\label{fig:onecol}
\end{figure*}

There are two branches. In the spatial relation branch, for each object $i$,  a polar coordinate system is defined centered at $i$, the 4-dimension bounding box position $p_j=(x,y,w,h)$ of object $j$ is transferred into a polar coordinate vector $(\rho_j,\theta_j)$.  Because it is quite efficient to express spatial orientation between centers of the bounding boxes $i$ and $j$. In descriptions of images, few absolute position is used to describe the spatial relationships of objects in an image. On the contrary, relative position is widely used (\textit{e.g. 'on', 'below' and so on}. Our method focus on relative position of objects rather than absolute position. It means that we pay more attention to "What are the neighborhoods around a particular object and where are they?". The relative distances and angles of objects' center can indicate spatial relation intuitively.

Low-dimensional relative positions are embedded to higher dimension through a set of Gaussian kernels\cite{Monti_2017_CVPR} with learnable means and  covariances of Gaussian distributions, where the spatial relation of between objects $i$ and $j$ are easy separable. The spatial relation dimension after embedding is $d_p=64$ experimentally. The kernel operator for object $j$ centered at $i$ is defined as follows:
\begin{equation}\label{equa:2}
\begin{aligned}
\omega_{\rho_j}=exp(-\frac{\parallel \rho_j - \rho_0 \parallel_2^2}{2\sigma_{\rho}^2})
\end{aligned} \end{equation}
\begin{equation}\label{equa:2}
\begin{aligned}
\omega_{\theta_j}=exp(-\frac{\parallel \theta_j - \theta_0 \parallel_2^2}{2\sigma_{\theta}^2})
\end{aligned} \end{equation}
where $\rho_0$, $\sigma_{\rho}$ are the learnable means and covariances of Gaussian distributions for relevant distance and $\theta_0$, $\sigma_{\theta}$ are the learnable means and covariances of Gaussian distributions for relevant angle.

We aggregates the relative distance and angle relationship of objects $i$ and $j$ with a scaling function, which means that the strength of spatial relationships between objects can be weighted by spatial orientation. The aggregated spatial weight for image is represented as $\omega_p$.
\begin{equation}\label{equa:2}
\begin{aligned}
\omega_p=\frac{\omega_{\rho_j} \omega_{\theta_j}}{\sum_{j=1}^N\omega_{\rho_j} \omega_{\theta_j}}
\end{aligned} \end{equation}
where $N$ is the number of objects in an image.

In another branch, image features $V$ is linearly transformed by $f(\cdot)$. Semantic relation $\omega_s$ is computed as in Eq.4. Dot-product attention is employed in our algorithm with a scaling factor $\frac{1}{\sqrt{d_v}}$, where $d_v$ is the dimension of object features $v$.
 \begin{equation}\label{equa:3}
\begin{aligned}
\omega_s=\frac{f(V)^T f(V)}{\sqrt{d_v}}
\end{aligned} \end{equation}
where $d_v$ is the dimension of object feature $V$.

The intra-image relation weight $\omega_I$ indicates both the semantic and spatial impact from object $j$.
Spatial relationship $\omega_p$ of different objects is fused with semantic relationship $\omega_s$ between objects through Eq.5. It is scaled in the range (0, 1) and can be regarded as a variant of softmax. $\omega_I$ is computed as follows:

\begin{equation}\label{equa:4}
\begin{aligned}
\omega_I=\frac{\omega_p exp(\omega_s)}{\sum_{k=1}^{N}{\omega_p exp(\omega_s)}}
\end{aligned} \end{equation}

Multi-head attention \cite{NIPS2017_7181}is employed to adapt flexible relationships, since different heads can focus on different aspects of relation. Multiple relation features from multi heads are aggregated as follows:
 \begin{equation}\label{equa:5}
 \begin{aligned}
V_r= f(V) + Concat [(\omega_I f(V))_1,(\omega_I f(V))_2,...,(\omega_I f(V))_K]
 \end{aligned} \end{equation}
$K$ is the number of relation heads, which is typically set to be 6, same as transformer\cite{NIPS2017_7181}.

\subsection{Two-step Relation Network}
Image-text matching task aims to retrieve relevant images given a sentence query, and vice versa.

In order to capture interpretable alignment of image-text pairs. Inspired by the great effectiveness of attention mechanism, a two-step relation module is designed in our proposed ParNet.

In the first step, intra-modal relation mechanism computes responses between different entities in an image or a sentence separately. the image (sentence) representation is able to attend on the informative parts within each modality.

In the second step, the query plays a role of textual context to refine the relationship among image objects while inferring the similarity. Inter-modal relation mechanism balances the importance of each entities according to query's intent.
\subsubsection{Intra-modal relation module}
Intra-modal relation module acts as a self-attention mechanism to generate context-attended vectors by focusing on relevant parts of image and text separately.

The input of this module is object features $I=[V||P]$ extracted from Faster-RCNN and word vectors $T$ constructed by bi-directional GRU.

\textbf{Visual intra-modal relation module} For image, position-aware relation module is employed here, which aims to generate a context vector by focusing on certain relevant parts of the input image which is decided by intra-image relation weight $\Omega_I$.

An image can aware the accurate representation of objects in particular scenery through intra-modal relation matrix $\omega_I$ in Section 3.2, which contributes to image content understanding.

The encoded image representation is $V_r$ computed in Section 3.2.

\textbf{Textual intra-modal relation module} Like-wise, we apply an adaptive relation module to compute context features by focusing on relevant words of the input sentence. We apply scaled dot-product attention\cite{NIPS2017_7181} to compute weighted average vectors.

First, text features $T$ is linearly transformed by $g(\cdot)$.
The textual intra-modal relation weight matrix $\omega_T$ is computed as Eq.7 with a scaling factor $d_t$. Here $d_t$ is the dimension of word features $T$. A softmax function is applied to normalize intra-sentence weight matrix $\omega_T$.
 \begin{equation}\label{equa:3}
\begin{aligned}
\omega_T=softmax(\frac{g(T)^Tg(T)}{\sqrt{d_t}})
\end{aligned} \end{equation}

K-head relation features of input sentence is aggregated through sublayer addition. $K$ is set to be 6.
 \begin{equation}\label{equa:5}
 \begin{aligned}
T_r= g(T) + Concat [(\omega_T g(T))_1,(\omega_T g(T))_2,...,(\omega_T g(T))_K]
 \end{aligned} \end{equation}
\subsubsection{Inter-modal relation module}
For image-text matching, efficient and effective computation of cross-modal similarities is crucial. Given different queries, the importance of different image objects and the relevance between them are quite different.

Our inter-modal relation module considers the fact that the content of sentences have influence on the visual context. Sentences is considered as weak annotations for image understanding.

While inferring the similarity, we employ inter-modal relation module to compare each image object to  the corresponding sentence vector in order to determine the importance of image objects.

For each image-text pair, we focus on relevant objects in an image with respect to each word in the text.  Firstly, we measure the object-word relevance as weight matrix as follows:
\begin{equation}\label{equa:3}
\begin{aligned}
u_{ij}=\frac{{v_r}_i \cdot {t_r}_i}{\parallel {v_r}_i\parallel \parallel {t_r}_j \parallel}
\end{aligned} \end{equation}
 \begin{equation}\label{equa:4}
\begin{aligned}
\alpha_i=\frac{1}{M}\sum_{j=1}^M softmax(\lambda \cdot u_{ij}) v_r^i
\end{aligned} \end{equation}
Where ${v_r}_i$ represents $i-th$ image object and ${t_r}_j$ represents $j-th$ word in Section 3.3.1.    $u_{ij}$ is the inter-modal relevance between $i-th$ object and $j-th$ word. The output of this module $\{\alpha_1,\alpha_2,...,\alpha_N\}$ is a set of image objects under the constraint of corresponding sentence annotation, where N is the number of image objects.

The importance distribution of image objects are balanced according to corresponding sentence's intent. The object-word alignments inferred after this step are more comprehensive.
\subsection{Similarity Alignments}
we compute the cosine similarity between each attended image and the corresponding sentence to obtain retrieval scores of input image-text pair.
 \begin{equation}\label{equa:3}
\begin{aligned}
s(i,j)=\frac{\alpha_i \cdot {t_r}_j}{\parallel \alpha_i \parallel\parallel {t_r}_j\parallel}
\end{aligned} \end{equation}

In addition,Both self-relation module and cross-relation module discover all possible alignments simultaneously, which is quite time efficient.
The final similarity can be summarized with average pooling(AVG):
 \begin{equation}\label{equa:4}
\begin{aligned}
S(I,T)=\frac{\sum_{j=1}^{M}\sum_{i=1}^{N}s(i,j)}{N \cdot M}
\end{aligned} \end{equation}

%daixiugai

Our method is trained with bidirectional max-margain ranking loss, which is widely used in multi-modal matching tasks. For each positive image-text pair (I,T), negative image-text pairs $(I,T^-)$ and $(I^-,T)$ are sampled.In practice, for computational efficiency, rather than summing over all the negative samples, it usually considers only the hard negatives in a mini-batch of stochastic gradient descent.The network is trained to minimize the distance of positive pairs while maximizing that of negative pairs. The Loss function is defined as Eq.18.

 \begin{equation}\label{equa:4}
\begin{aligned}
\mathcal{L} (I,T)=max(0,\beta-S[I,T]+S(I,T^-))\\
+max(0,\beta-S(I,T)+S(I^-,T))
\end{aligned} \end{equation}
Where $\beta$ is a margin constraint. By minimizing this function, our network can focus on the important entities which appears in correct image-text pairs through self-relation and cross-attention modules.

\begin{figure*}
\begin{center}
\includegraphics[width=0.90\linewidth]{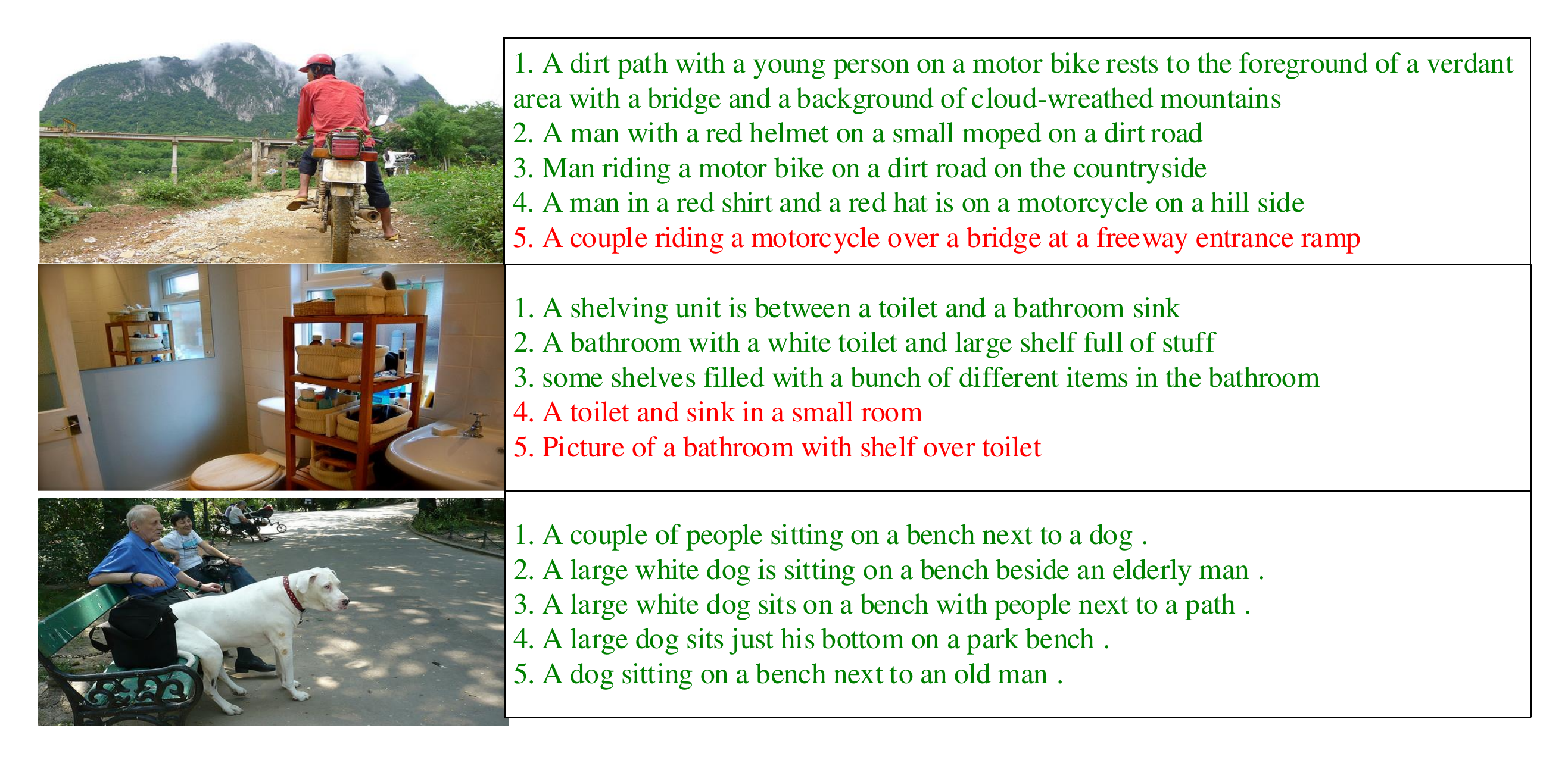}
\end{center}
   \caption{The qualitative results of image-to-text retrieval. The first column is image queries, the second column shows top-5 retrieved results. Green for correct answers, and red for wrong answers}
\label{fig:long}
\label{fig:onecol}
\end{figure*}

\begin{figure*}[htb]
\begin{center}
\includegraphics[width=0.95\linewidth]{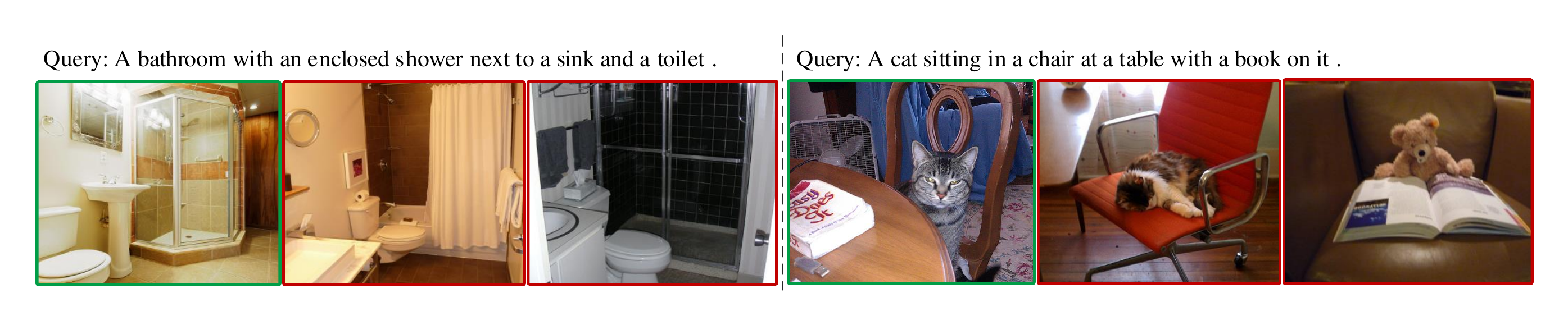}
\end{center}
   \caption{The qualitative results of text-to-image retrieval.The first line is text queries, the second line shows top-3 retrieved results. Green for correct answer, and red for wrong answers.}
\label{fig:long}
\label{fig:onecol}
\end{figure*}

\section{Experiments}
\subsection{Datasets and Evaluation}
\subsubsection{Datasets}
We evaluate our image-text matching model the Microsoft COCO dataset\cite{DBLP:journals/corr/LinMBHPRDZ14}.
%and Flickr30k dataset.
Microsoft COCO dataset contains 123287 images with five descriptive sentences for each. We use the same spilt as \cite{Karpathy_2015_CVPR}, 1000 images are used for validation , 1000 for testing. In \cite{Karpathy_2015_CVPR}, 82783 images are spilt into training set, 5000 for validation and 5000 for test. Following \cite{Faghri2017VSEIV}, 30504 images are added into training set that were originally in the validation set to improve accuracy. Totally,  there are 113287 images in the training set. Our results are reported by either averaging over 5 folds of 1K test images or testing on the full 5K test images.
% Flickr 30k contains 31000 real images. We follow the public spilts by: 29000 training, validation and 1000 testing images. Each image is annotated with five corresponding sentences.

% \begin{table}
%   \caption{Frequency of Special Characters}
%   \label{tab:freq}
%   \begin{tabular}{ccl}
%     \toprule
%     Non-English or Math&Frequency&Comments\\
%     \midrule
%     \O & 1 in 1,000& For Swedish names\\
%     $\pi$ & 1 in 5& Common in math\\
%     \$ & 4 in 5 & Used in business\\
%     $\Psi^2_1$ & 1 in 40,000& Unexplained usage\\
%   \bottomrule
% \end{tabular}
% \end{table}

\subsubsection{Evaluation Metrics}
We adopt Recall@K (K=1,5,10) for retrieval evaluation.
Recall@K represents the percentage of the queries where at least one ground-truth is retrieved among the top K results.

\subsection{Implementation Details}
The whole system is trained by Adam optimizer\cite{Kingma2015AdamAM}. Models are trained for 25 epochs. The initial learning rate is set as 0.0005. We use a mini-batch size of 128 for all experiments. Margin constraint $\beta$ is set as 0.2.

\subsection{Results on MS-COCO}

% \begin{table*}
% \begin{center}
% \caption{Comparision of the image-text matching results on MS-COCO dataset in terms of Recall@K(R@K)}
% %\begin{tabular}{|c|p{1cm}<{\centering} p{1cm}<{\centering} p{1cm}<{\centering}|p{1cm}<{\centering} p{1cm}<{\centering} p{1cm}<{\centering}|}
% \begin{tabular}{c|c c c c|c c c c}
% \toprule
% \multirow{2}*{Methods} & \multicolumn{4}{c}{Image-to-Text} & \multicolumn{4}{c}{Text-to-Image} \\
% \cline{2-9} & R@1 & R@5 & R@10 & Med $r$ & R@1 & R@5 & R@10 &Med $r$\\
% \midrule
% DVSA \cite{Karpathy_2015_CVPR} & 38.4 &69.9 &80.5&1.0 &27.4 & 60.2& 74.8&3.0 \\
% mCNN\cite{DBLP:journals/corr/MaLSL15} & 42.8 &73.1 &84.1 &2.0&32.6 &68.6 & 82.8&3.0 \\
% HM-LSTM\cite{Niu_2017_ICCV} & 43.9 &- &87.8 &2.0 & 36.1 & - &86.7 &3.0\\
% DSPE\cite{DBLP:journals/corr/WangLL15} &50.1 &89.2 &- &-&39.6& 86.9& -& -\\
% VSE++\cite{Faghri2017VSEIV} & 64.6 &- &95.7&1.0& 52.0& - &92.0&1.0 \\
% DPC \cite{DBLP:journals/corr/abs-1711-05535} & 65.6 &89.8 &95.5&1 &47.1 &79.9 &90.0 &2\\
% Gen-GXN  \cite{Gu_2018_CVPR} & 68.5 &- &97.9&1 &56.6 &- &94.5&1 \\
% SCO  \cite{Yan2017Learning} & 69.9 &92.9 &97.5&-& 56.7 &87.5 &\textbf{94.8} &-\\
% SCAN \cite{Lee_2018_ECCV} &70.9 &94.5 &97.8 &-&56.4 &87.0 &93.9 &-\\
% \midrule

% ParNet (NP)&72.8&\textbf{94.9}&97.9&1.0&57.9&87.4&94&1.0\\
% ParNet (P)&\textbf{73.5}&94.5&\textbf{98.3}&1.0&\textbf{58.3}&\textbf{88.2}&94.1&1.0\\
% \bottomrule
% \end{tabular}
% \end{center}
% \end{table*}

\begin{table}
\begin{center}
\caption{Comparision of the image-text matching results on MS-COCO dataset in terms of Recall@K(R@K)}
%\begin{tabular}{|c|p{1cm}<{\centering} p{1cm}<{\centering} p{1cm}<{\centering}|p{1cm}<{\centering} p{1cm}<{\centering} p{1cm}<{\centering}|}
\begin{tabular}{c|c c c|c c c}
\toprule
\multirow{2}*{Methods} & \multicolumn{3}{c}{Image-to-Text} & \multicolumn{3}{c}{Text-to-Image} \\
\cline{2-7} & R@1 & R@5 & R@10  & R@1 & R@5 & R@10 \\
\midrule
DVSA \cite{Karpathy_2015_CVPR} & 38.4 &69.9 &80.5 &27.4 & 60.2& 74.8 \\
mCNN\cite{DBLP:journals/corr/MaLSL15} & 42.8 &73.1 &84.1&32.6 &68.6 & 82.8 \\
HM-LSTM\cite{Niu_2017_ICCV} & 43.9 &- &87.8  & 36.1 & - &86.7\\
DSPE\cite{DBLP:journals/corr/WangLL15} &50.1 &89.2 &-&39.6& 86.9& -\\
VSE++\cite{Faghri2017VSEIV} & 64.6 &- &95.7& 52.0& - &92.0\\
DPC \cite{DBLP:journals/corr/abs-1711-05535} & 65.6 &89.8 &95.5 &47.1 &79.9 &90.0\\
Gen-GXN  \cite{Gu_2018_CVPR} & 68.5 &- &97.9 &56.6 &- &94.5 \\
SCO  \cite{Yan2017Learning} & 69.9 &92.9 &97.5& 56.7 &87.5 &\textbf{94.8} \\
SCAN \cite{Lee_2018_ECCV} &70.9 &94.5 &97.8&56.4 &87.0 &93.9\\
\midrule

ParNet (NP)&72.8&\textbf{94.9}&97.9&57.9&87.4&94\\
ParNet (P)&\textbf{73.5}&94.5&\textbf{98.3}&\textbf{58.3}&\textbf{88.2}&94.1\\
\bottomrule
\end{tabular}
\end{center}
\end{table}

\begin{table}[hbt]
\begin{center}
\caption{Ablation study of ParNet for image-text matching results on MS-COCO dataset}
\setlength{\tabcolsep}{0.8mm}{
\begin{tabular}{l|c c c|c c c}
\toprule
\multirow{2}*{Methods} & \multicolumn{3}{c}{Image-to-Text} & \multicolumn{3}{c}{Text-to-Image} \\
\cline{2-7}
& R@1 & R@5 & R@10 & R@1 & R@5 & R@10 \\
\midrule
base  &70.9 &94.5 &97.8 &56.4 &87.0 &93.9 \\
\midrule
+intra-modal relation(only T) &72.3&95.3&98.2&57.5&87.1&93.8\\
+intra-modal relation(only I) &71.8&95.1&97.8&58.0&87.5&93.9\\
\midrule
+intra \& inter modal relation &72.8&\textbf{94.9}&97.9&57.9&87.4&94.0\\
\midrule
+ position ($d_p$=32)&69.3&94.8&98.4&56.6&87.6&94.0\\
+ position ($d_p$=64) &73.5&94.5&98.3&58.3&88.2&94.1\\
+ position ($d_p$=128)&71.8&95.2&98.2&58.7&87.6&93.9\\
\midrule
ParNet &\textbf{73.5}&94.5&\textbf{98.3}&\textbf{58.3}&\textbf{88.2}&\textbf{94.1}\\
\bottomrule
\end{tabular}}
\end{center}
\end{table}

% \begin{figure}[h]
% \begin{center}
% \includegraphics[width=0.99\linewidth]{badcase.pdf}
% \end{center}
%   \caption{Bad case of i2t retrieval results.}
% \label{fig:long}
% \label{fig:onecol}
% \end{figure}

\begin{figure*}[h]
\begin{center}
\includegraphics[width=0.90\linewidth]{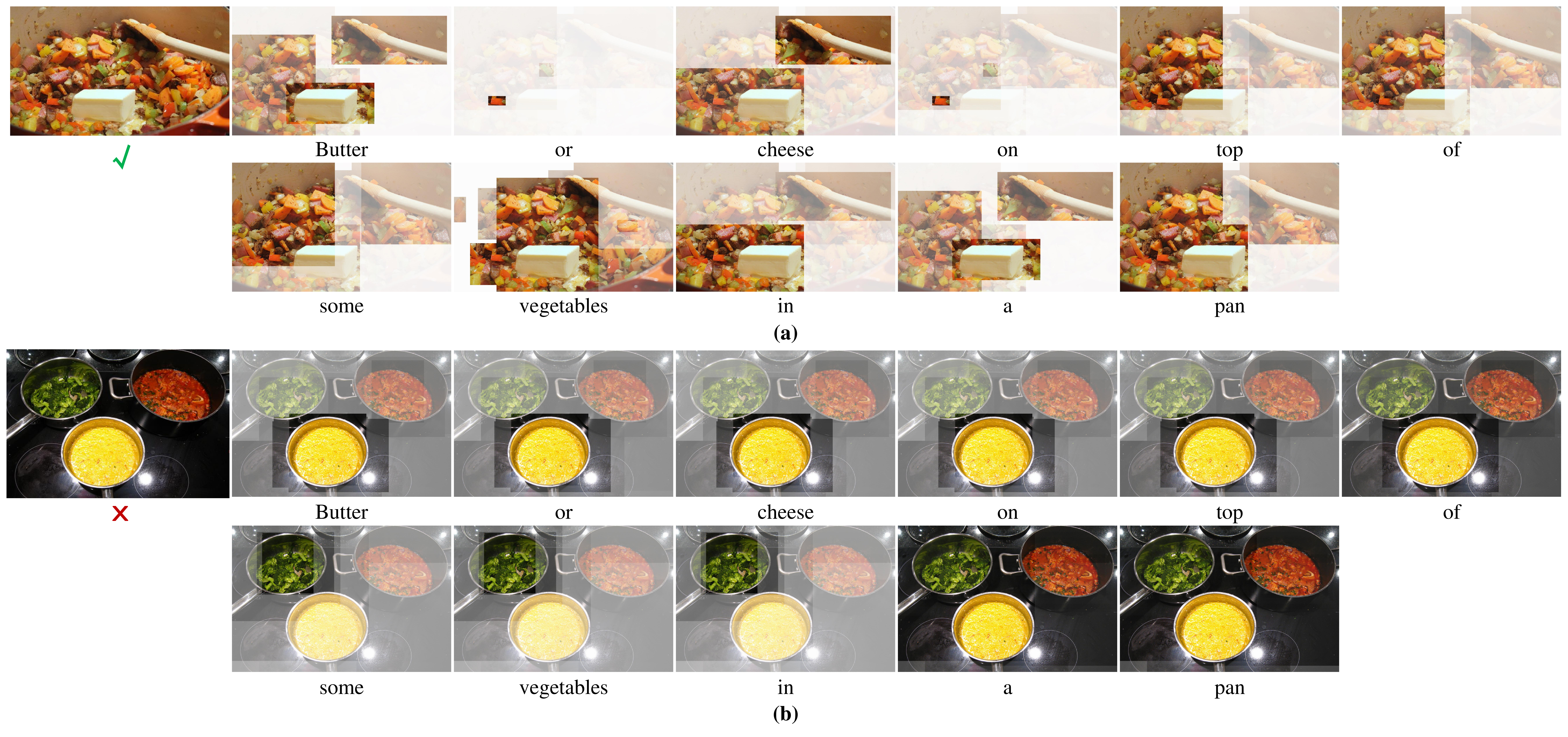}
\end{center}
   \caption{The visualization of attention outputs in the second step relation of text-to-image retrieval for query "Butter or cheese on \textit{top} of some vegetables \textit{in} a pan. " (a) is the top-1 retrieval results of ParNet with position-aware model. It's right. (b) is the top-1 retrieval results of ParNet without position-aware module. It's wrong. While (a) and (b) can both detect entities corresponding nouns, \textit{e.g.}, 'vegetables','pan'. The result of ParNet without position-aware module fails to focus on the relative position,  \textit{e.g., on}, and returns a wrong answer.}
\label{fig:long}
\label{fig:onecol}
\end{figure*}

We compare our methods with recently developed methods on the image-to-text and text-to-image retrieval tasks. The quantitative performance on MS-COCO test set of our method compared with state-of-the-art methods is presented in Table 1.
The compared methods are as follows:

\textbf{DVSA} \cite{Karpathy_2015_CVPR} develops a deep neural network model that infers the latent alignment between segments of sentences and the objects of images, it is a combination of CNN and bidirectional RNN.

\textbf{mCNN}\cite{DBLP:journals/corr/MaLSL15} consists of one image CNN encoding the image content and one matching CNN modeling the joint representation of image and sentence.

\textbf{HM-LSTM}\cite{Niu_2017_ICCV} exploits the hierarchical relations between sentences and phrases, and between whole images and image objects, to jointly establish their representations.

\textbf{DSPE}\cite{DBLP:journals/corr/WangLL15} uses a two-branch neural network with multiple layers of linear projections followed by nonlinearities to learn joint embeddings of image-text pairs.

\textbf{VSE++}\cite{Faghri2017VSEIV} embeds whole images and sentences to a common space without the use of attention mechanism and also leveraged hard negatives sampling.

\textbf{DPC} \cite{DBLP:journals/corr/abs-1711-05535} proposes a instance loss which explicitly considers the intra-modal data distribution based on an unsupervised assumption that each image / text group can be viewed as a class.

\textbf{Gen-GXN}  \cite{Gu_2018_CVPR} incorporates generative processes into the cross-modal feature embedding to learn global abstract features and local grounded features simultaneous.

\textbf{SCO}  \cite{Yan2017Learning} improves the image representation by learning semantic concepts and then organizing them in a correct semantic order.

\textbf{SCAN} \cite{Lee_2018_ECCV}  considers the fact that the importance of words can depend on the visual context and employ cross attention to discover possible alignments.

For fair evaluation, we compare single model accuracy obtained without model ensemble or fused.
We can see that our ParNet outperforms previous approaches on MS-COCO dataset. R@1 achieves 73.5\% for image-to-text retrieval, which is improved by 3.67\%  compared with SCAN (single model)\cite{Lee_2018_ECCV}. R@1 for text-to-image retrieval achieves 58.3\%, 2.84\% higher than SCAN.
The improvement results of our method shows the effectiveness of inferring the latent relationships.

In Table 2, our method without position-aware relation module achieves the result of R@1=72.8\%, which is superior to our baseline. It confirms the effectiveness of our aggregated two-step relation module. Our methods ParNet with position-aware module achieves 73.5\% for R@1. It is obvious that the added spatial information help capturing the correspondence between image and text.

The qualitative results of image-to-text retrieval and text-to-image retrieval are illustrated in Figure 4 and Figure 5 respectively. In Figure 4, the first column is image queries, the second column is the top-5 retrieved results. The green ones represent for correct answers, and red ones represent for wrong answers.  We can see that, some of the wrong answers occurred in the results also have similar meaning with the corresponding query(\textit{e.g.}, 'A toliet and sink in a small room.'). In Figure 5 (b), the third results capture the spatial relation \textit{'in'} correctly, but it mistake teddy bear for cat.

\subsection{Visualization analyse of positon-aware module}

The visualization of attention outputs in the second step relation of text-to-image retrieval for query "Butter or cheese on \textit{top} of some vegetables \textit{in} a pan. " Figure 7 (a) is the top-1 retrieval results of ParNet with position-aware model. It's right.  Figure 6 (b) is the top-1 retrieval results of ParNet without position-aware module. It's wrong. While Figure 6 (a) and  Figure 6 (b) can both detect entities corresponding nouns, \textit{e.g.}, 'vegetables','pan'. The result of ParNet without position-aware relation module fails to focus on the relative position,  \textit{e.g., on}, and returns a wrong answer. It confirms the effectiveness of position-aware relation module.
And the correct correspondence between attended objects and words (\textit{e.g.}, 'butter', 'vegetables','pan') in the query shows that our two-step aggregated relation module is quite available for capturing latent image-text alignments between different modalities.

\section{Conclusion}
In this paper, we propose a position-aware aggregated relation network for bridging the semantic gaps of image and text. We propose an position-aware relation module for image to obtain both the semantic and spatial relationship. Then we combine intra-modal relation and inter-modal relation for capturing the alignments of entities in images (or sentences). This model achieves the state-of-the-art performance for image-text matching task, showing the effectiveness in extracting latent alignments between image-text pairs.

\bibliographystyle{ACM-Reference-Format}
\bibliography{PAARN}


\begin{thebibliography}{30}

%%% the Web address.

\ifx \showCODEN    \undefined \def \showCODEN     #1{\unskip}     \fi
\ifx \showDOI      \undefined \def \showDOI       #1{#1}\fi
\ifx \showISBNx    \undefined \def \showISBNx     #1{\unskip}     \fi
\ifx \showISBNxiii \undefined \def \showISBNxiii  #1{\unskip}     \fi
\ifx \showISSN     \undefined \def \showISSN      #1{\unskip}     \fi
\ifx \showLCCN     \undefined \def \showLCCN      #1{\unskip}     \fi
\ifx \shownote     \undefined \def \shownote      #1{#1}          \fi
\ifx \showarticletitle \undefined \def \showarticletitle #1{#1}   \fi
\ifx \showURL      \undefined \def \showURL       {\relax}        \fi
% invisible to TeX
\providecommand\bibfield[2]{#2}
\providecommand\bibinfo[2]{#2}
\providecommand\natexlab[1]{#1}
\providecommand\showeprint[2][]{arXiv:#2}

\bibitem[\protect\citeauthoryear{Anderson, He, Buehler, Teney, Johnson, Gould,
  and Zhang}{Anderson et~al\mbox{.}}{2018}]%
        {Anderson_2018_CVPR}
\bibfield{author}{\bibinfo{person}{Peter Anderson}, \bibinfo{person}{Xiaodong
  He}, \bibinfo{person}{Chris Buehler}, \bibinfo{person}{Damien Teney},
  \bibinfo{person}{Mark Johnson}, \bibinfo{person}{Stephen Gould}, {and}
  \bibinfo{person}{Lei Zhang}.} \bibinfo{year}{2018}\natexlab{}.
\newblock \showarticletitle{Bottom-Up and Top-Down Attention for Image
  Captioning and Visual Question Answering}. In \bibinfo{booktitle}{\emph{The
  IEEE Conference on Computer Vision and Pattern Recognition (CVPR)}}.
\newblock


\bibitem[\protect\citeauthoryear{Dehghani, Gouws, Vinyals, Uszkoreit, and
  Kaiser}{Dehghani et~al\mbox{.}}{2018}]%
        {DBLP:journals/corr/abs-1807-03819}
\bibfield{author}{\bibinfo{person}{Mostafa Dehghani}, \bibinfo{person}{Stephan
  Gouws}, \bibinfo{person}{Oriol Vinyals}, \bibinfo{person}{Jakob Uszkoreit},
  {and} \bibinfo{person}{Lukasz Kaiser}.} \bibinfo{year}{2018}\natexlab{}.
\newblock \showarticletitle{Universal Transformers}.
\newblock \bibinfo{journal}{\emph{CoRR}}  \bibinfo{volume}{abs/1807.03819}
  (\bibinfo{year}{2018}).
\newblock
\showeprint[arxiv]{1807.03819}
\urldef\tempurl%
\url{http://arxiv.org/abs/1807.03819}
\showURL{%
\tempurl}


\bibitem[\protect\citeauthoryear{Devlin, Chang, Lee, and Toutanova}{Devlin
  et~al\mbox{.}}{2018}]%
        {DBLP:journals/corr/abs-1810-04805}
\bibfield{author}{\bibinfo{person}{Jacob Devlin}, \bibinfo{person}{Ming{-}Wei
  Chang}, \bibinfo{person}{Kenton Lee}, {and} \bibinfo{person}{Kristina
  Toutanova}.} \bibinfo{year}{2018}\natexlab{}.
\newblock \showarticletitle{{BERT:} Pre-training of Deep Bidirectional
  Transformers for Language Understanding}.
\newblock \bibinfo{journal}{\emph{CoRR}}  \bibinfo{volume}{abs/1810.04805}
  (\bibinfo{year}{2018}).
\newblock
\showeprint[arxiv]{1810.04805}
\urldef\tempurl%
\url{http://arxiv.org/abs/1810.04805}
\showURL{%
\tempurl}


\bibitem[\protect\citeauthoryear{Dong, Li, and Snoek}{Dong
  et~al\mbox{.}}{2016}]%
        {Dong2016Word2VisualVec}
\bibfield{author}{\bibinfo{person}{Jianfeng Dong}, \bibinfo{person}{Xirong Li},
  {and} \bibinfo{person}{Cees G.~M. Snoek}.} \bibinfo{year}{2016}\natexlab{}.
\newblock \showarticletitle{Word2VisualVec: Cross-Media Retrieval by Visual
  Feature Prediction}.
\newblock  (\bibinfo{year}{2016}).
\newblock


\bibitem[\protect\citeauthoryear{Faghri, Fleet, Kiros, and Fidler}{Faghri
  et~al\mbox{.}}{2017}]%
        {Faghri2017VSEIV}
\bibfield{author}{\bibinfo{person}{Fartash Faghri}, \bibinfo{person}{David~J.
  Fleet}, \bibinfo{person}{Ryan Kiros}, {and} \bibinfo{person}{Sanja Fidler}.}
  \bibinfo{year}{2017}\natexlab{}.
\newblock \showarticletitle{VSE++: Improved Visual-Semantic Embeddings}.
\newblock \bibinfo{journal}{\emph{CoRR}}  \bibinfo{volume}{abs/1707.05612}
  (\bibinfo{year}{2017}).
\newblock


\bibitem[\protect\citeauthoryear{Frome, Corrado, Shlens, Bengio, Dean, Ranzato,
  and Mikolov}{Frome et~al\mbox{.}}{2013}]%
        {NIPS2013_5204}
\bibfield{author}{\bibinfo{person}{Andrea Frome}, \bibinfo{person}{Greg~S
  Corrado}, \bibinfo{person}{Jon Shlens}, \bibinfo{person}{Samy Bengio},
  \bibinfo{person}{Jeff Dean}, \bibinfo{person}{Marc\textquotesingle~Aurelio
  Ranzato}, {and} \bibinfo{person}{Tomas Mikolov}.}
  \bibinfo{year}{2013}\natexlab{}.
\newblock \showarticletitle{DeViSE: A Deep Visual-Semantic Embedding Model}.
\newblock In \bibinfo{booktitle}{\emph{Advances in Neural Information
  Processing Systems 26}}, \bibfield{editor}{\bibinfo{person}{C.~J.~C. Burges},
  \bibinfo{person}{L.~Bottou}, \bibinfo{person}{M.~Welling},
  \bibinfo{person}{Z.~Ghahramani}, {and} \bibinfo{person}{K.~Q. Weinberger}}
  (Eds.). \bibinfo{publisher}{Curran Associates, Inc.},
  \bibinfo{pages}{2121--2129}.
\newblock
\urldef\tempurl%
\url{http://papers.nips.cc/paper/5204-devise-a-deep-visual-semantic-embedding-model.pdf}
\showURL{%
\tempurl}


\bibitem[\protect\citeauthoryear{Gu, Cai, Joty, Niu, and Wang}{Gu
  et~al\mbox{.}}{2018}]%
        {Gu_2018_CVPR}
\bibfield{author}{\bibinfo{person}{Jiuxiang Gu}, \bibinfo{person}{Jianfei Cai},
  \bibinfo{person}{Shafiq~R. Joty}, \bibinfo{person}{Li Niu}, {and}
  \bibinfo{person}{Gang Wang}.} \bibinfo{year}{2018}\natexlab{}.
\newblock \showarticletitle{Look, Imagine and Match: Improving Textual-Visual
  Cross-Modal Retrieval With Generative Models}. In
  \bibinfo{booktitle}{\emph{The IEEE Conference on Computer Vision and Pattern
  Recognition (CVPR)}}.
\newblock


\bibitem[\protect\citeauthoryear{Hardoon, Szedmak, and Shawe-Taylor}{Hardoon
  et~al\mbox{.}}{2004}]%
        {Hardoon2004Canonical}
\bibfield{author}{\bibinfo{person}{David~R. Hardoon},
  \bibinfo{person}{Sandor~R. Szedmak}, {and} \bibinfo{person}{John~R.
  Shawe-Taylor}.} \bibinfo{year}{2004}\natexlab{}.
\newblock \bibinfo{booktitle}{\emph{Canonical Correlation Analysis: An Overview
  with Application to Learning Methods}}.
\newblock \bibinfo{publisher}{MIT Press}.
\newblock


\bibitem[\protect\citeauthoryear{Hu, Gu, Zhang, Dai, and Wei}{Hu
  et~al\mbox{.}}{2018}]%
        {Hu_2018_CVPR}
\bibfield{author}{\bibinfo{person}{Han Hu}, \bibinfo{person}{Jiayuan Gu},
  \bibinfo{person}{Zheng Zhang}, \bibinfo{person}{Jifeng Dai}, {and}
  \bibinfo{person}{Yichen Wei}.} \bibinfo{year}{2018}\natexlab{}.
\newblock \showarticletitle{Relation Networks for Object Detection}. In
  \bibinfo{booktitle}{\emph{The IEEE Conference on Computer Vision and Pattern
  Recognition (CVPR)}}.
\newblock


\bibitem[\protect\citeauthoryear{Karpathy and Fei-Fei}{Karpathy and
  Fei-Fei}{2015}]%
        {Karpathy_2015_CVPR}
\bibfield{author}{\bibinfo{person}{Andrej Karpathy} {and} \bibinfo{person}{Li
  Fei-Fei}.} \bibinfo{year}{2015}\natexlab{}.
\newblock \showarticletitle{Deep Visual-Semantic Alignments for Generating
  Image Descriptions}. In \bibinfo{booktitle}{\emph{The IEEE Conference on
  Computer Vision and Pattern Recognition (CVPR)}}.
\newblock


\bibitem[\protect\citeauthoryear{Karpathy, Joulin, and Li}{Karpathy
  et~al\mbox{.}}{2014}]%
        {KarpathyJF14}
\bibfield{author}{\bibinfo{person}{Andrej Karpathy}, \bibinfo{person}{Armand
  Joulin}, {and} \bibinfo{person}{Fei{-}Fei Li}.}
  \bibinfo{year}{2014}\natexlab{}.
\newblock \showarticletitle{Deep Fragment Embeddings for Bidirectional Image
  Sentence Mapping}.
\newblock \bibinfo{journal}{\emph{CoRR}}  \bibinfo{volume}{abs/1406.5679}
  (\bibinfo{year}{2014}).
\newblock
\showeprint[arxiv]{1406.5679}
\urldef\tempurl%
\url{http://arxiv.org/abs/1406.5679}
\showURL{%
\tempurl}


\bibitem[\protect\citeauthoryear{Kingma and Ba}{Kingma and Ba}{2015}]%
        {Kingma2015AdamAM}
\bibfield{author}{\bibinfo{person}{Diederik~P. Kingma} {and}
  \bibinfo{person}{Jimmy Ba}.} \bibinfo{year}{2015}\natexlab{}.
\newblock \showarticletitle{Adam: A Method for Stochastic Optimization}.
\newblock \bibinfo{journal}{\emph{CoRR}}  \bibinfo{volume}{abs/1412.6980}
  (\bibinfo{year}{2015}).
\newblock


\bibitem[\protect\citeauthoryear{Krishna, Zhu, Groth, Johnson, Hata, Kravitz,
  Chen, Kalantidis, Li, Shamma, Bernstein, and Li}{Krishna
  et~al\mbox{.}}{2016}]%
        {DBLP:journals/corr/KrishnaZGJHKCKL16}
\bibfield{author}{\bibinfo{person}{Ranjay Krishna}, \bibinfo{person}{Yuke Zhu},
  \bibinfo{person}{Oliver Groth}, \bibinfo{person}{Justin Johnson},
  \bibinfo{person}{Kenji Hata}, \bibinfo{person}{Joshua Kravitz},
  \bibinfo{person}{Stephanie Chen}, \bibinfo{person}{Yannis Kalantidis},
  \bibinfo{person}{Li{-}Jia Li}, \bibinfo{person}{David~A. Shamma},
  \bibinfo{person}{Michael~S. Bernstein}, {and} \bibinfo{person}{Fei{-}Fei
  Li}.} \bibinfo{year}{2016}\natexlab{}.
\newblock \showarticletitle{Visual Genome: Connecting Language and Vision Using
  Crowdsourced Dense Image Annotations}.
\newblock \bibinfo{journal}{\emph{CoRR}}  \bibinfo{volume}{abs/1602.07332}
  (\bibinfo{year}{2016}).
\newblock
\showeprint[arxiv]{1602.07332}
\urldef\tempurl%
\url{http://arxiv.org/abs/1602.07332}
\showURL{%
\tempurl}


\bibitem[\protect\citeauthoryear{Lee, Chen, Hua, Hu, and He}{Lee
  et~al\mbox{.}}{2018}]%
        {Lee_2018_ECCV}
\bibfield{author}{\bibinfo{person}{Kuang-Huei Lee}, \bibinfo{person}{Xi Chen},
  \bibinfo{person}{Gang Hua}, \bibinfo{person}{Houdong Hu}, {and}
  \bibinfo{person}{Xiaodong He}.} \bibinfo{year}{2018}\natexlab{}.
\newblock \showarticletitle{Stacked Cross Attention for Image-Text Matching}.
  In \bibinfo{booktitle}{\emph{The European Conference on Computer Vision
  (ECCV)}}.
\newblock


\bibitem[\protect\citeauthoryear{Lin, Maire, Belongie, Bourdev, Girshick, Hays,
  Perona, Ramanan, Doll{\'{a}}r, and Zitnick}{Lin et~al\mbox{.}}{2014}]%
        {DBLP:journals/corr/LinMBHPRDZ14}
\bibfield{author}{\bibinfo{person}{Tsung{-}Yi Lin}, \bibinfo{person}{Michael
  Maire}, \bibinfo{person}{Serge~J. Belongie}, \bibinfo{person}{Lubomir~D.
  Bourdev}, \bibinfo{person}{Ross~B. Girshick}, \bibinfo{person}{James Hays},
  \bibinfo{person}{Pietro Perona}, \bibinfo{person}{Deva Ramanan},
  \bibinfo{person}{Piotr Doll{\'{a}}r}, {and} \bibinfo{person}{C.~Lawrence
  Zitnick}.} \bibinfo{year}{2014}\natexlab{}.
\newblock \showarticletitle{Microsoft {COCO:} Common Objects in Context}.
\newblock \bibinfo{journal}{\emph{CoRR}}  \bibinfo{volume}{abs/1405.0312}
  (\bibinfo{year}{2014}).
\newblock
\showeprint[arxiv]{1405.0312}
\urldef\tempurl%
\url{http://arxiv.org/abs/1405.0312}
\showURL{%
\tempurl}


\bibitem[\protect\citeauthoryear{Ma, Lu, Shang, and Li}{Ma
  et~al\mbox{.}}{2015}]%
        {DBLP:journals/corr/MaLSL15}
\bibfield{author}{\bibinfo{person}{Lin Ma}, \bibinfo{person}{Zhengdong Lu},
  \bibinfo{person}{Lifeng Shang}, {and} \bibinfo{person}{Hang Li}.}
  \bibinfo{year}{2015}\natexlab{}.
\newblock \showarticletitle{Multimodal Convolutional Neural Networks for
  Matching Image and Sentence}.
\newblock \bibinfo{journal}{\emph{CoRR}}  \bibinfo{volume}{abs/1504.06063}
  (\bibinfo{year}{2015}).
\newblock
\showeprint[arxiv]{1504.06063}
\urldef\tempurl%
\url{http://arxiv.org/abs/1504.06063}
\showURL{%
\tempurl}


\bibitem[\protect\citeauthoryear{Monti, Boscaini, Masci, Rodola, Svoboda, and
  Bronstein}{Monti et~al\mbox{.}}{2017}]%
        {Monti_2017_CVPR}
\bibfield{author}{\bibinfo{person}{Federico Monti}, \bibinfo{person}{Davide
  Boscaini}, \bibinfo{person}{Jonathan Masci}, \bibinfo{person}{Emanuele
  Rodola}, \bibinfo{person}{Jan Svoboda}, {and} \bibinfo{person}{Michael~M.
  Bronstein}.} \bibinfo{year}{2017}\natexlab{}.
\newblock \showarticletitle{Geometric Deep Learning on Graphs and Manifolds
  Using Mixture Model CNNs}. In \bibinfo{booktitle}{\emph{The IEEE Conference
  on Computer Vision and Pattern Recognition (CVPR)}}.
\newblock


\bibitem[\protect\citeauthoryear{Nam, Ha, and Kim}{Nam et~al\mbox{.}}{2017}]%
        {Nam_2017_CVPR}
\bibfield{author}{\bibinfo{person}{Hyeonseob Nam}, \bibinfo{person}{Jung-Woo
  Ha}, {and} \bibinfo{person}{Jeonghee Kim}.} \bibinfo{year}{2017}\natexlab{}.
\newblock \showarticletitle{Dual Attention Networks for Multimodal Reasoning
  and Matching}. In \bibinfo{booktitle}{\emph{The IEEE Conference on Computer
  Vision and Pattern Recognition (CVPR)}}.
\newblock


\bibitem[\protect\citeauthoryear{Niu, Zhou, Wang, Gao, and Hua}{Niu
  et~al\mbox{.}}{2017}]%
        {Niu_2017_ICCV}
\bibfield{author}{\bibinfo{person}{Zhenxing Niu}, \bibinfo{person}{Mo Zhou},
  \bibinfo{person}{Le Wang}, \bibinfo{person}{Xinbo Gao}, {and}
  \bibinfo{person}{Gang Hua}.} \bibinfo{year}{2017}\natexlab{}.
\newblock \showarticletitle{Hierarchical Multimodal LSTM for Dense
  Visual-Semantic Embedding}. In \bibinfo{booktitle}{\emph{The IEEE
  International Conference on Computer Vision (ICCV)}}.
\newblock


\bibitem[\protect\citeauthoryear{Ren, He, Girshick, and Sun}{Ren
  et~al\mbox{.}}{2015}]%
        {DBLP:journals/corr/RenHG015}
\bibfield{author}{\bibinfo{person}{Shaoqing Ren}, \bibinfo{person}{Kaiming He},
  \bibinfo{person}{Ross~B. Girshick}, {and} \bibinfo{person}{Jian Sun}.}
  \bibinfo{year}{2015}\natexlab{}.
\newblock \showarticletitle{Faster {R-CNN:} Towards Real-Time Object Detection
  with Region Proposal Networks}.
\newblock \bibinfo{journal}{\emph{CoRR}}  \bibinfo{volume}{abs/1506.01497}
  (\bibinfo{year}{2015}).
\newblock
\showeprint[arxiv]{1506.01497}
\urldef\tempurl%
\url{http://arxiv.org/abs/1506.01497}
\showURL{%
\tempurl}


\bibitem[\protect\citeauthoryear{Scholkopf and Smola}{Scholkopf and
  Smola}{2001}]%
        {Scholkopf:2001:LKS:559923}
\bibfield{author}{\bibinfo{person}{Bernhard Scholkopf} {and}
  \bibinfo{person}{Alexander~J. Smola}.} \bibinfo{year}{2001}\natexlab{}.
\newblock \bibinfo{booktitle}{\emph{Learning with Kernels: Support Vector
  Machines, Regularization, Optimization, and Beyond}}.
\newblock \bibinfo{publisher}{MIT Press}, \bibinfo{address}{Cambridge, MA,
  USA}.
\newblock
\showISBNx{0262194759}


\bibitem[\protect\citeauthoryear{{Schuster} and {Paliwal}}{{Schuster} and
  {Paliwal}}{1997}]%
        {650093}
\bibfield{author}{\bibinfo{person}{M. {Schuster}} {and} \bibinfo{person}{K.~K.
  {Paliwal}}.} \bibinfo{year}{1997}\natexlab{}.
\newblock \showarticletitle{Bidirectional recurrent neural networks}.
\newblock \bibinfo{journal}{\emph{IEEE Transactions on Signal Processing}}
  \bibinfo{volume}{45}, \bibinfo{number}{11} (\bibinfo{date}{Nov}
  \bibinfo{year}{1997}), \bibinfo{pages}{2673--2681}.
\newblock
\showISSN{1053-587X}
\urldef\tempurl%
\url{https://doi.org/10.1109/78.650093}
\showDOI{\tempurl}


\bibitem[\protect\citeauthoryear{Simonyan and Zisserman}{Simonyan and
  Zisserman}{2014}]%
        {article}
\bibfield{author}{\bibinfo{person}{Karen Simonyan} {and}
  \bibinfo{person}{Andrew Zisserman}.} \bibinfo{year}{2014}\natexlab{}.
\newblock \showarticletitle{Very Deep Convolutional Networks for Large-Scale
  Image Recognition}.
\newblock \bibinfo{journal}{\emph{arXiv 1409.1556}} (\bibinfo{date}{09}
  \bibinfo{year}{2014}).
\newblock


\bibitem[\protect\citeauthoryear{Tsai, Huang, and Salakhutdinov}{Tsai
  et~al\mbox{.}}{2017}]%
        {DBLP:journals/corr/TsaiHS17}
\bibfield{author}{\bibinfo{person}{Yao{-}Hung~Hubert Tsai},
  \bibinfo{person}{Liang{-}Kang Huang}, {and} \bibinfo{person}{Ruslan
  Salakhutdinov}.} \bibinfo{year}{2017}\natexlab{}.
\newblock \showarticletitle{Learning Robust Visual-Semantic Embeddings}.
\newblock \bibinfo{journal}{\emph{CoRR}}  \bibinfo{volume}{abs/1703.05908}
  (\bibinfo{year}{2017}).
\newblock
\showeprint[arxiv]{1703.05908}
\urldef\tempurl%
\url{http://arxiv.org/abs/1703.05908}
\showURL{%
\tempurl}


\bibitem[\protect\citeauthoryear{Vaswani, Shazeer, Parmar, Uszkoreit, Jones,
  Gomez, Kaiser, and Polosukhin}{Vaswani et~al\mbox{.}}{2017}]%
        {NIPS2017_7181}
\bibfield{author}{\bibinfo{person}{Ashish Vaswani}, \bibinfo{person}{Noam
  Shazeer}, \bibinfo{person}{Niki Parmar}, \bibinfo{person}{Jakob Uszkoreit},
  \bibinfo{person}{Llion Jones}, \bibinfo{person}{Aidan~N Gomez},
  \bibinfo{person}{\L~ukasz Kaiser}, {and} \bibinfo{person}{Illia Polosukhin}.}
  \bibinfo{year}{2017}\natexlab{}.
\newblock \showarticletitle{Attention is All you Need}.
\newblock In \bibinfo{booktitle}{\emph{Advances in Neural Information
  Processing Systems 30}}, \bibfield{editor}{\bibinfo{person}{I.~Guyon},
  \bibinfo{person}{U.~V. Luxburg}, \bibinfo{person}{S.~Bengio},
  \bibinfo{person}{H.~Wallach}, \bibinfo{person}{R.~Fergus},
  \bibinfo{person}{S.~Vishwanathan}, {and} \bibinfo{person}{R.~Garnett}}
  (Eds.). \bibinfo{publisher}{Curran Associates, Inc.},
  \bibinfo{pages}{5998--6008}.
\newblock
\urldef\tempurl%
\url{http://papers.nips.cc/paper/7181-attention-is-all-you-need.pdf}
\showURL{%
\tempurl}


\bibitem[\protect\citeauthoryear{Wang, Yang, Xu, Hanjalic, and Shen}{Wang
  et~al\mbox{.}}{2017b}]%
        {Wang2017Adversarial}
\bibfield{author}{\bibinfo{person}{Bokun Wang}, \bibinfo{person}{Yang Yang},
  \bibinfo{person}{Xing Xu}, \bibinfo{person}{Alan Hanjalic}, {and}
  \bibinfo{person}{Heng~Tao Shen}.} \bibinfo{year}{2017}\natexlab{b}.
\newblock \showarticletitle{Adversarial Cross-Modal Retrieval}. In
  \bibinfo{booktitle}{\emph{ACM on Multimedia Conference}}.
  \bibinfo{pages}{154--162}.
\newblock


\bibitem[\protect\citeauthoryear{Wang, Li, and Lazebnik}{Wang
  et~al\mbox{.}}{2015}]%
        {DBLP:journals/corr/WangLL15}
\bibfield{author}{\bibinfo{person}{Liwei Wang}, \bibinfo{person}{Yin Li}, {and}
  \bibinfo{person}{Svetlana Lazebnik}.} \bibinfo{year}{2015}\natexlab{}.
\newblock \showarticletitle{Learning Deep Structure-Preserving Image-Text
  Embeddings}.
\newblock \bibinfo{journal}{\emph{CoRR}}  \bibinfo{volume}{abs/1511.06078}
  (\bibinfo{year}{2015}).
\newblock
\showeprint[arxiv]{1511.06078}
\urldef\tempurl%
\url{http://arxiv.org/abs/1511.06078}
\showURL{%
\tempurl}


\bibitem[\protect\citeauthoryear{Wang, Girshick, Gupta, and He}{Wang
  et~al\mbox{.}}{2017a}]%
        {DBLP:journals/corr/abs-1711-07971}
\bibfield{author}{\bibinfo{person}{Xiaolong Wang}, \bibinfo{person}{Ross~B.
  Girshick}, \bibinfo{person}{Abhinav Gupta}, {and} \bibinfo{person}{Kaiming
  He}.} \bibinfo{year}{2017}\natexlab{a}.
\newblock \showarticletitle{Non-local Neural Networks}.
\newblock \bibinfo{journal}{\emph{CoRR}}  \bibinfo{volume}{abs/1711.07971}
  (\bibinfo{year}{2017}).
\newblock
\showeprint[arxiv]{1711.07971}
\urldef\tempurl%
\url{http://arxiv.org/abs/1711.07971}
\showURL{%
\tempurl}


\bibitem[\protect\citeauthoryear{Yan, Qi, and Liang}{Yan et~al\mbox{.}}{2017}]%
        {Yan2017Learning}
\bibfield{author}{\bibinfo{person}{Huang Yan}, \bibinfo{person}{Wu Qi}, {and}
  \bibinfo{person}{Wang Liang}.} \bibinfo{year}{2017}\natexlab{}.
\newblock \showarticletitle{Learning Semantic Concepts and Order for Image and
  Sentence Matching}.
\newblock \bibinfo{journal}{\emph{IEEE Transactions on Pattern Analysis and
  Machine Intelligence}} (\bibinfo{year}{2017}).
\newblock


\bibitem[\protect\citeauthoryear{Zheng, Zheng, Garrett, Yang, and Shen}{Zheng
  et~al\mbox{.}}{2017}]%
        {DBLP:journals/corr/abs-1711-05535}
\bibfield{author}{\bibinfo{person}{Zhedong Zheng}, \bibinfo{person}{Liang
  Zheng}, \bibinfo{person}{Michael Garrett}, \bibinfo{person}{Yi Yang}, {and}
  \bibinfo{person}{Yi{-}Dong Shen}.} \bibinfo{year}{2017}\natexlab{}.
\newblock \showarticletitle{Dual-Path Convolutional Image-Text Embedding}.
\newblock \bibinfo{journal}{\emph{CoRR}}  \bibinfo{volume}{abs/1711.05535}
  (\bibinfo{year}{2017}).
\newblock
\showeprint[arxiv]{1711.05535}
\urldef\tempurl%
\url{http://arxiv.org/abs/1711.05535}
\showURL{%
\tempurl}


\end{thebibliography}
\end{document}